\documentclass[10pt,twocolumn,letterpaper]{article}

\usepackage{cvpr}
\usepackage{times}
\usepackage{epsfig}
\usepackage{graphicx}
\usepackage{amsmath}
\usepackage{amssymb}

\usepackage[tight,scriptsize]{subfigure}
\usepackage{multirow}

\newcommand{\be}{\begin{equation}}
\newcommand{\ee}{\end{equation}}
\def\comment#1{{{\color{red}}}}
\def\newcomment#1{{{#1}}}
\def\cut#1{{{\color{green}}}}
\def\newcut#1{{{\color{cyan}}}}
\def\change#1{{{\color{red} #1}}}
\def\change#1{{{#1}}}
\def\finalchange#1{{{#1}}}

\usepackage[breaklinks=true,colorlinks,bookmarks=false]{hyperref}

\cvprfinalcopy 

\def\cvprPaperID{326} 

\ifcvprfinal\fi
\begin{document}

\title{Visual-Inertial-Semantic Scene Representation for 3D Object Detection}

\author{Jingming Dong\thanks{Equal contributors.}~~~~~~~~~~Xiaohan Fei$^*$~~~~~~~~~~Stefano Soatto\\
UCLA Vision Lab, University of California, Los Angeles, CA 90095\\
{\tt\small \{dong, feixh, soatto\}@cs.ucla.edu}
}

\maketitle

\begin{abstract}
We describe a system to detect objects in three-dimensional space using video and inertial sensors (accelerometer and gyrometer), ubiquitous in modern mobile platforms from phones to drones. Inertials afford the ability to impose class-specific scale priors for objects, and provide a global orientation reference. A minimal sufficient representation, the posterior of semantic (identity) and syntactic (pose) attributes of objects in space, can be decomposed into a geometric term, which can be maintained by a localization-and-mapping filter, and a likelihood function, which can be approximated by a discriminatively-trained convolutional neural network. \cut{Co-occurrence priors can be enforced to rebalance the evidence.} The resulting system can process the video stream causally in real time, and provides a representation of objects in the scene that is persistent: Confidence in the presence of objects grows with evidence, and objects previously seen are kept in memory even when temporarily occluded, with their return into view automatically predicted to prime re-detection.
\end{abstract}

\section{Introduction}
We deem an ``object detector'' to be a system that takes as input {\em images} and produces as output decisions as to the presence of {\em objects in the scene.} We design one based on the following premises: (a) Objects exist in the scene, not in the image; (b) they persist, so confidence on their presence should grow as more evidence is accrued from multiple (test) images; (c) once seen, the system should be aware of their presence even when temporarily not visible; (d) such awareness should allow it to predict when they will return into view, based on scene geometry and topology; \newcut{(e) awareness of everything else in the scene (context) should allow the system to predict when objects may come into view, before they are even visible, or at least to rebalance the evidence based on co-occurrence statistics; }(e) objects have characteristic shape and {\em size} in 3D, and vestibular (inertial) sensors provide a global scale and orientation reference that the system should leverage on.

Detecting objects from images is not the same as detecting images of objects (Fig.~\ref{fig-toy-car})\cut{, which is all one can do given {\em a single (test) image}}. Objects do not flicker in-and-out of existence, and do not disappear when not seen (Fig.~\ref{fig-occ}). What we call ``object detectors'' traditionally refers to algorithms that process a single image and return a decision as to the presence of objects of a certain class in said image, missing several critical elements (a)-(e) above. Nevertheless, such algorithms can be modified to produce {\em not} decisions, but {\em evidence} (likelihood) for the presence of objects, which can be processed over time and integrated against the geometric and topological structure of the {\em scene}, to yield an object detector that has the desired characteristics. The scene context encompasses both the identity and co-occurrence of objects (semantics) but also their spatial arrangement in three-dimensional (3D) space (syntax). 

\begin{figure}[t]
\begin{center}
\subfigure{\label{fig-1a}\includegraphics[width=.94\columnwidth]{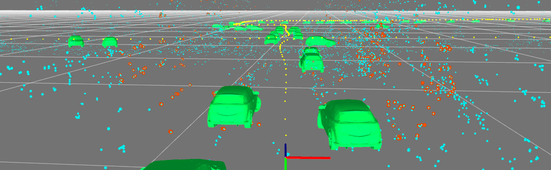}}\vspace{-.3cm}
\subfigure{\label{fig-1b}\includegraphics[width=.94\columnwidth]{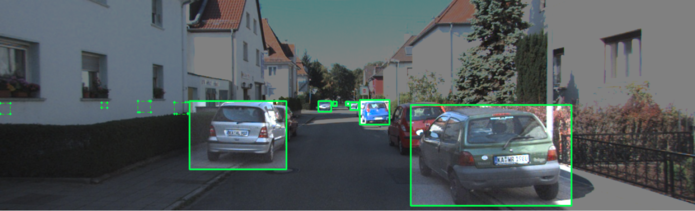}}\vspace{-.3cm}
\subfigure{\label{fig-1c}\includegraphics[width=.94\columnwidth]{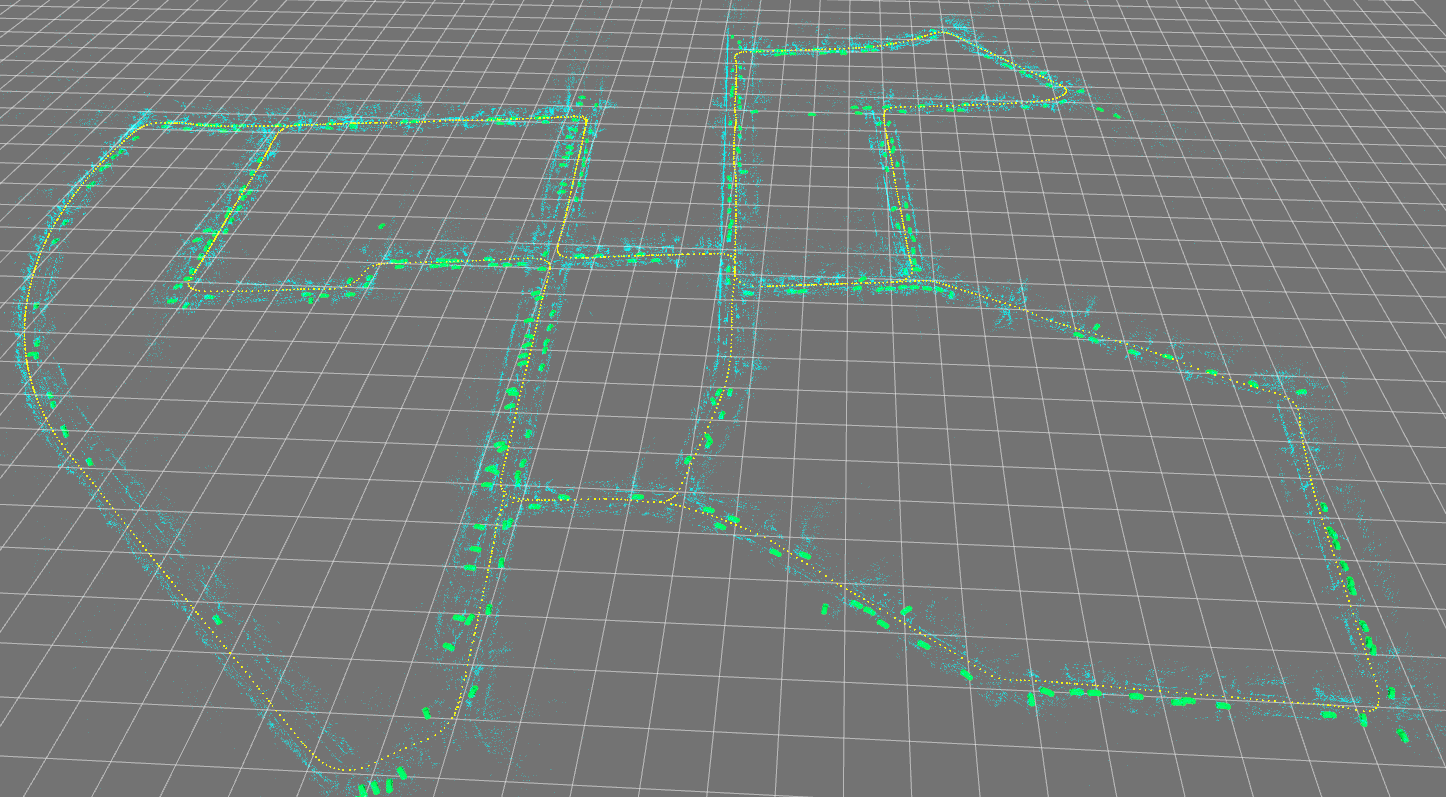}}
\end{center}
 \caption{\comment{middle image too dark}{\sl Illustration of our system to detect objects-in-scenes.} Top: state of the system with reconstructed scene representation (cyan), currently tracked points (red), viewer trajectory from a previous loop (yellow) and current pose (reference frame). All cars detected are shown as point-estimates (the best-aligned generic CAD model) in green, including those previously-seen on side streets (far left). Middle: visualization of the implicit measurement process: Objects in the state are projected onto the current image based on the mean vehicle pose estimate (green boxes) and their likelihood score is computed (visualized as contrast: sharp regions have high likelihood, dim regions low). Cars in different streets, known to not be visible, are visualized as dashed boxes and their score discarded. Bottom: Top view of the state from the entire KITTI-00 sequence (best viewed at $5\times$)\newcut{, best viewed at $5\times$}. \newcut{Additional representative experiments are shown in the Sup.~Mat.  \comment{DRAW IN DASHED LINES, SWAP FIRST AND SECOND, AND DIM IMAGE OUTSIDE BOUNDING BOXES}} }
\label{fig-teaser}
\end{figure}

\subsection{Summary of Contributions and Limitations}

To design an object detector based on the premises above, we (a) formalize an explicit model of the posterior probability of object attributes, both semantic (identity) and syntactic (pose), natively in the 3D scene (Sect. \ref{sec:formalization}), which (b) maintains and updates such a posterior, processing each image causally over time (Sect.~\ref{sec:approximations}); (c) the posterior distribution is a form of short-term memory (representation), which we use to (d) predict visibility and occlusion relations (Sect. \ref{sec:occlusion}). \newcut{We briefly describe (e) a simple re-weighting of the evidence based on a co-occurrence prior (Sect. \ref{sect-prior}).} We exploit the availability of cheap inertial sensors in almost every mobile computing platform to (e) impose class-specific priors on the size of objects (Sect. \ref{sec:scale}).

The key insight from the formalization (a) above is that an optimal (minimal sufficient invariant \cite{soattoC16ICLR}) representation for objects in the scene (Eq.~\ref{eq-evol}) can be factored into two components: One geometric -- which can be computed recursively by a localization (SLAM) system (Eq.~\ref{eq-slam}) -- and the other a likelihood term, which can be evaluated instantaneously by a discriminatively-trained convolutional neural network (CNN, Eq.~\ref{eq-approx})  operating on a single image. Some consequences of this insight are discussed in Sect. \ref{sec:discussion}. In practice, this means that we can implement our system using some off-the-shelf components, fine-tuning a pre-trained CNN, and at least for some rudimentary modeling assumptions, our system operates in real-time, generating object-scene representations at 10-30 frames per second. In Sect.~\ref{sec:experiments} we report the results of a representative sample of qualitative and quantitative tests.

Our system is the first to exploit inertial sensors to provide both scale discrimination and global orientation for visual recognition (Fig.~\ref{fig-toy-car}). Most (image)-object detectors assume images are gravity-aligned, which is a safe bet for photographic images, not so for robots or drones. Our system is also the first to integrate CNN-based detectors in a recursive Bayesian inference scheme, and to implement the overall system to run in real-time \cite{cvpr-demo-anonymous}.

While our formalization of the problem of object detection is general, our real-time implementation has several limitations. First, it only returns a joint geometric and semantic description for {\em static objects}. Moving objects are detected in the image, but their geometry -- shape and pose, estimating which would require sophisticated class-specific deformation priors -- is not inferred. Second, it models objects' shape as a parallelepiped, or bounding box in 3D. While this is a step forward from bounding boxes in the image, it is still a rudimentary model of objects, based on which visibility computation is rather crude. We have performed several tests with dense reconstruction \cite{graberal15}, as well as with CAD models \cite{izadinia2016im2cad}, but matching and visibility computation based on those is not yet at the level of accuracy (dense reconstruction) or efficiency (CAD matching) to enable real-time computation. 
The third limitation is that a full joint syntactic-semantic prior is not enforced. While ideally we would like to predict not only what objects are likely to become visible based on context, but also {\em where} they will appear relative to each other, this is still computationally prohibitive at scale. \newcut{In our case, we only use a co-occurrence prior \cite{choi2010exploiting} to rebalance the likelihood, so that unlikely objects (\eg, motorcycles and boats) that otherwise would show as detected in an image, are discounted based on the context.}

In Sect.~\ref{sec:formalization} we start by defining an object representation as a sufficient invariant for detection, and show that the main factor can be updated recursively as an integral, where the measure represents the syntactic context, and can be computed by a SLAM system, and the other factor can be computed by a CNN. While the update is straightforward and {\em top-down} (the system state generate predictions for image-projections, whose likelihood is scored by a CNN), initialization requires defining a prior on object identity and pose. For this we use the same CNN in a {\em bottom-up} mode, where putative detection (high-likelihood regions) are used to initialize object hypotheses (or, rather, regions with no putative detections are assumed free of objects), and several heuristics are put in place for genetic phenomena (birth, death and merging of objects, Sect. \ref{sec:implementation}). 

\cut{In Sect. \ref{sec:expm}, we report several qualitative experiments on several representative datasets, including indoor and outdoor. More importantly, our real-time implementation has already been publicly demonstrated in a live demo \cite{cvpr-demo-anonymous}. Generating quantitative comparisons with competing approaches is non-trivial, since we are not aware of any other method that exploits both visual and inertial sensing, nor of any dataset that provides such data for benchmarking object detection system. There are (very few) datasets for visual and inertial sensor fusion, but they do not have ground-truth object annotations. Similarly, there are many object detection datasets, some in 3D, but they do not provide time-stamped inertial measurements.
The closest benchmark dataset we can find in the public domain is KITTI \cite{kitti}. While inertial data is not provided, stereo can be used for scale reference. Annotation is sparse, so this is not an extensive test, but we report results for all the objects and frames that are provided with ground-truth. To facilitate progress in this area, we plan to make our datasets, which include validated trajectories in 3D and time-stamped inertial and full-frame VGA camera measurements, public upon the completion of the anonymous review process, along with our implementation, for  reproducibility.
}

\section{Related Work}
\label{sect-relatedwork}

This work, by its nature, relates to a vast body of literature on scene understanding in Computer Vision, Robotics \cite{laiBRF12,pillai_rss15}  and AI \cite{koppula2011semantic} dating back decades \cite{waltz1981understanding}. Most recently, with the advent of cheap consumer range sensors, there has been a wealth of activity in this area \cite{lin2013holistic,toshev2012shape,wu2014hierarchical,cadena2013semantic,singh2013nonparametric,deng2015semantic,gupta2013perceptual,karpathy2013object,salas2013slam++,hermans2014dense,banica2015second,sengupta2013,vineet2015incremental, kehl2016deep,song2016deep,ren2016three}. The use of RGB-D cameras unfortunately restricts the domain of applicability mostly indoors and at close range whereas we target mobility applications where the camera, which typically has an inertial sensor strapped on it, but not (yet) a range sensor, can be used both indoor and outdoors. We expect that, on indoor sequences, our method would underperform a structured light or other RGB-D source, but this is subject of future investigation. 

There is also work that focuses on scene understanding from visual sensors, specifically video \cite{kundu2014joint,aditya2015visual,lin2016exploring,silberman2012indoor,baraldi2015scene,yao2012describing}, although none integrates inertial data, despite a resurgent interest in sensor fusion \cite{zhang2015sensor}. Additional related work includes \cite{hane2013joint,cornelis20083d,brostow2008segmentation,savinov2015discrete}.

To the best of our knowledge, no work leverages inertial sensing for object detection. This is critical to provide a scale estimate in a monocular setting, and validate object hypotheses in a Bayesian setting, so that, for instance, a model car in our system is not classified as a car (Fig. \ref{fig-toy-car}). 

Semantic scene understanding {\em from a single image} is also an area of research (\cite{fouhey2015single} and references therein). We are instead interested in agents embedded in physical space, for which the restriction to a single image is limiting. There is also a vast literature on scene segmentation (\cite{hoiem2015guest} and references therein), mostly using range (RGB-D) sensors.  
{One popular pipeline for dense semantic segmentation is adopted by \cite{hermans2014dense,mccormac2016semanticfusion,vineet2015incremental, kundu2014joint,asif2016simultaneous}: Depth maps obtained either from RGB-D or stereo are fused; 2D semantic labeling is transferred to 3D and smoothed with a fully-connected CRF~\cite{koltun2011efficient}. Also related methods on joint semantic segmentation and reconstruction are \cite{savinov2016semantic,ulusoypatches,blahalarge}.}

There is also work on 3D recognition \cite{kim20133d,sharma2014recursive,mottaghi2014role}, but again with no inertial measurements and no motion. Some focus on real-time operation \cite{couprie2014convolutional}, but most operate off-line \cite{zia2015towards,chhayamonocular}. None of the datasets commonly used in these works \cite{cordts2016cityscapes,xiao2013sun3d} provide an inertial reference, except for KITTI. In terms of 3D object detection on KITTI, some authors focus on image-based detection \cite{girshick2014rich,girshick2015fast,ren2015faster,redmon2016you,liu2016ssd} and then place objects into the scene \cite{3dvp,subcnn}, while others focus on 3D object proposal generation and verification using a network \cite{chen2016monocular,chen20153d}.  
\cite{3dvp} trains a 3D Voxel Pattern (3DVP) based detector to infer object attributes and demonstrates the ability to accurately localize cars in 3D on KITTI. Their subsequent work \cite{subcnn} trains a CNN to classify 3DVPs. Different representations of object proposals are also exploited, such as 3D cuboids \cite{fidler20123d} and deformable 3D wireframes \cite{zia2015towards}. Various priors are also considered: \cite{wang2015holistic} exploits geo-tagged images; geometric priors of objects are incorporated into various optimization frameworks to estimate object attributes \cite{zhu2015single,chhayamonocular}. While most of these algorithms report very good performance on detection ($\sim90\%$ mean average precision), none reports scores for the semantic-syntactic state of objects in 3D, except for \cite{3dvp,subcnn} and \cite{chen20153d, chen2016monocular}. Since the latter are dominated by the former, we take \cite{subcnn} as a paragon for comparison in Sect.~\ref{sec:experiments}.

The aforementioned 3D object recognition methods are based on 2D detection without temporal consistency. Therefore, the comparison is somewhat unfair as single-image based detectors cannot reliably detect objects in space, which is our main motivation for the proposed approach. For details on comparison methodology, see Sect.~\ref{sec:experiments}. \change{\cite{chhayamonocular, song2015joint} use multiple views, but their output is a point-estimate instead of a posterior. Also, the optimization has to be re-run once new datum is available.}

Recent work in data association \cite{leonardos2016distributed} aims to directly infer the association map, which is computationally prohibitive for the scale needed in our real-time system. We therefore resort to heuristics, described in Sect. \ref{sec:implementation}. More specifically to our implementation, we leverage existing visual-inertial filters \cite{hesch2013towards,li2014online,tsotsosCS15} and single image-trained CNNs \cite{girshick2014rich,redmon2016you,subcnn}. 



\section{Methods}
\label{sec:formalization}


\def\M{l} 
\def\S{s} 
\def\X{{x}} 
\def\I{I} 
\def\g{g} 
\def\u{u} 
\def\phii{\phi} 
\def\on#1{_{|_{#1}}}
\def\h{h} 
\def\rhoo{\rho} 
\def\k{\kappa} 
\def\x{{x_{i}}} 
\def\n{n} 
\def\y{y} 
\def\z{z}
\def\xii{\xi}
\def\zi{\z}

\subsection{Representations}

A scene $\xii$ is populated by a number of objects $\zi_j \in \{\zi_1,  \dots, \zi_N\}$, each with geometric (pose, shape)\footnote{Object pose is its position and orientation in world frame. With inertials, pose can be reduced to position and rotation around gravity. Sensor pose is full 6 degree-of-freedom position and orientation.} and semantic (label) attributes \newcomment{$\zi_j = \{\S_j, \M_j\}$}.\cut{\footnote{A scene instance can thus be thought of as a graph with nodes $\zi^j = \{\zi_1, \dots, \zi_j\}$ and edges capturing (geometric, positional and semantic) relations.}} Measurements ({\em e.g.,} images) up to the current time $t$, $\y^t \doteq \{\y_1, \dots, \y_t\}$ are captured from a sensor at pose $\g_t$. A {\em semantic} representation of the scene is the joint posterior $p(\xii, \zi^j | \y^t)$ for up to the $j$-th objects seen up to time $t$, where sensor pose $\g_t$ and other nuisances are marginalized. The joint posterior can be decomposed\cut{\footnote{If the goal is to infer objects $\zi$ from images $\y^t$, marginalizing the above would ignore the context $\xii$. This would make it impossible to exploit context knowledge while learning objects directly from data. However, the context model can be learned from object knowledge (first factor), and used to validate object hypotheses (second factor). Assuming a mixture model for $p(\xii | \zi)$, we can then exploit the context model while inferring objects via $\max_\xii p(\zi, \xii | \y^t)$. Alternatively, an empirical Bayes approach would first infer $\hat \xii = \arg\max p(\xii | \y^t)$ and then perform inference on $\zi$ using $p(\zi | \hat \xii, \y^t)$. We adopt the former method for simplicity.}}
 as 
$
p(\xii, \z^j | \y^t) = {p(\xii | \zi^j)}{p(\zi^j | \y^t)}
$
with the first factor ideally updated asynchronously each time a new object $\zi_{j+1}$ becomes manifest starting from a prior $p(\xii)$ 
and the second factor \newcut{\eqref{eq-evol}}updated each time a new measurement $\y_{t+1}$ becomes available starting from $t = 0$ and given $p(\zi)$. 

A representation of the scene in support of {\em (geometric) localization tasks} is the posterior $p(\g_t, \X | \y^t)$ over sensor pose $\g_t$ (which, of course, is not a nuisance for this task) and a sparse attributed\footnote{Attributes include sparse geometry (position in the inertial frame) and local photometry (feature descriptor, sufficient for local correspondence).} point cloud $\X = [\X_1, \dots, \X_{{}_{N_\X}}]$, given all measurements (visual $I^t$ and inertial $u^t$) up to the current time. Conditioning the semantics on the geometry\cut{\footnote{Note that these approximations allow us to separate the computation of the geometric representation (the ``dorsal stream'' to determine ``where'') from the semantic representation (the ``ventral stream'' to determine ``what'').}} we can write the second factor above as
\be
p(\zi^j | \y^t) = \int p(\zi^j | \g_t, \X, \y^t)
dP(\g_t, \X | \y^t) 
\label{eq-evol}
\ee
where the integrand can be updated as more data $y_{t+1}$ becomes available as $p(\zi^j | \g_{t+1}, \X, \y^{t+1})$, which is proportional to 
\be
p(\y_{t+1} | \zi^j, \g_{t+1}, \X)
\int p(\g_{t+1} | \g_t, \u_t) dP(\zi^j | \g_t, \X,  \y^t).
\label{eq-evol2}
\ee

\subsection{Approximations}
\label{sec:approximations}

The measure in \eqref{eq-evol} can be approximated in wide-sense using an Extended Kalman Filter (EKF), as customary in simultaneous localization and mapping (SLAM):  $p(\g_t, \X | \y^t) \simeq {\cal N}(\hat \g_{t|t}, \hat \X_{t|t};  P_{t|t})$. \eqref{eq-evol} is a diffusion around the mean/mode $\hat \g_{t|t}, \hat \X_{t|t}$; if the covariance $P_{t|t}$ is small, it can be further approximated: Given
\be
\hat \g_{t|t}, \hat \X_{t|t} = \arg\max_{\g_t, \X} p_{{}_{\rm SLAM}}(\g_t, \X | \y^t),
\label{eq-slam}
\ee
$\hat p_{\g, \X}(\zi^j | \y^t) \doteq p(\zi^j | \g_t = \hat \g_{t|t}, \X =  \hat \X_{t|t}, \y^t) \simeq 
p(\zi^j | \y^t)
$.
Otherwise the marginalization in \eqref{eq-evol} can be performed using samples from the SLAM system. Either way, omitting the subscripts, we have
\be
\hat p(\zi | \y^{t+1}) \propto \underbrace{p(\y_{t+1} | \zi, \hat g_{t|t}u_t, \hat \X_{t|t})}_{\rm CNN} \underbrace{\hat p(\zi | \y^t)}_{\rm BF}
\label{eq-approx}
\ee
where the likelihood term is approximated by a convolutional neural network (CNN) as shown in Sect.~\ref{sect-meas} and the posterior is updated by a Bayesian filter (BF) approximated by a bank of EKFs (Sect. \ref{sect-dependencies}). \newcomment{That only leaves the first factor $p(\xii | \zi^j)$ in the posterior, which encodes context. While one could approximate it with a recurrent network, that would be beyond our scope here; we even forgo using the co-occurrence prior, which amounts to a matrix multiplication that rebalances the classes following \cite{choi2010exploiting}, since for the limited number of classes and context priors we experimented with, it makes little difference.}

Approximating the likelihood in \eqref{eq-approx} appears daunting because of the purported need to generate future data $\y_{t+1}$ (the color of each pixel) from a given object class, shape  and pose, and to normalize with respect to all possible images of the object. Fortunately, the latter is not needed since the product on the right-hand side of \eqref{eq-approx} needs to be normalized anyway, which can be done easily in a particle/mixture-based representation of the posterior by dividing by the sum of the weights of the components. Generating actual images is similarly not needed. What is needed is a mechanism that, for a given image $\y_{t+1}$, allows quantifying the likelihood that an object of {\em any} class  with {\em any} shape  being present in {\em any} portion of the image  where it  projects to from the vantage point $\g_t$. In Sect. \ref{sect-meas} we will show how a discriminatively-trained CNN can be leveraged to this end.

\subsection{Measurement Process}
\label{sect-meas}
At each instant $t$, an image $\I_t$ is processed by ``probing functions'' $\phii$, which can be designed or trained to be invariant to nuisance variability. The SLAM system processes all past image measurements $\I^t$ and current inertial measurements $\u_t$, which collectively we refer to as $\y_t = \{\phii_\k(\I_t), \u_t\}$, where $\phii_\k(\I_t)$  is a collection of sparse contrast-invariant feature descriptors computed from the image for $N_i$ visible regions of the scene, and produces a joint posterior distribution of poses $\g_t$ and a sparse geometric representation of the scene $\X = [\X_1, \dots, \X_{N_i(t)}]$, assumed uni-modal and approximated by a Gaussian:
\be
p_{{}_{\rm SLAM}}(\g_t, \X | \y^t) \simeq {\cal N}(\hat \g_{t|t}, \hat \X_{t|t}; {P_{{}_{\{\g, \X\}}}}_{t|t})
\ee
where \newcomment{$\X \in \cup_j s_j$}, {\em i.e.,} the scene is assumed to be composed by the union of objects, including the default class ``background'' $\M_0$. This localization pipeline is borrowed from \cite{tsotsosCS15}, and is agnostic of the organization of the scene into objects and their identity. It also restricts $\X$ to a subset of the scene that is rigid, co-visible for a sufficiently long interval of time, and located on surfaces that, locally, exhibit Lambertian reflection.

To compute the marginal likelihood for each class $\M_k  \in \{\M_0, \dots, \M_K\}$, we leverage on a CNN trained discriminatively to classify a given image region $b_j$ into one of $K+1$ classes, including the background class. The architecture has a soft-max layer preceded by $K+1$ nodes, one per class, and is trained using the cross-entropy loss, providing a normalized score $\phii_{{}_{\rm CNN}}(\M | {\I_t}_{\on{b_j}})_{[k]}$ for each class and image bounding box $b_j$. We discard the soft-max layer, and forgo class-normalization.\newcut{and rebalance the classes based on their frequency in the training set} The activations at the $K+1$ nodes in the penultimate layer of the resulting network provide a mechanism for, given an image $\I_t$, quantifying the likelihood of each object class $\M_k$ being present at each bounding box $b_j$, which we interpret the  (marginal) likelihoods for (at least an instance of) each class being present at the given bounding box:
\be
\phii_{{}_{\rm CNN}}(\M | {\I_t}_{\on{b_j}})_{[k]} \simeq p(\I_t | \M_k, b_j).
\ee
 \cut{Note that the resulting CNN is not an object detector, but a likelihood function \cite{soattoC16ICLR}: It is not a discriminant for a random variable $\M$ that can take one of $K+1$ possible values, but instead a function of a $K+1$-dimensional (binary) vector $\M$, which is not normalized with respect to $\M$.}  
 \comment{INCONSISTENT NOTATION. where does $o$ go??Also $\xi$ does not appear from now on; THE NOTATION MUST BE CONSISTENT THROUGHOUT THE PAPER!!! EVERY SYMBOL MUST BE DEFINED BEFORE IT IS USED. SYMBOLS THAT ARE NEVER USED MUST BE REMOVED!} This process induces a likelihood on object classes being present \newcomment{in the {\em visible portion of the scene} regions of $\S_j$ and corresponding vantage points $\g_t$, via $b_j = \pi(\g_t \S_j)$ where $\pi$ is the projection.} Since inertials $u_t$ are directly measured, up to a Gaussian noise, we have: 
\be
p(y_t | z^j, \g_{t}, \change{x}) \simeq \phii_{{}_{\rm CNN}}(\M | {\I_t}_{\on{\pi(\g_t \S_j)}})_{[k]}{\cal N}(\bar u; Q)
\label{eq:cnn}
\ee
where $\bar u$ are the inertial biases and $Q$ the noise covariance; here the object attributes $z^j$ are the labels $l_j = l_k$ and \newcomment{geometry $s_j$}. Thus, given an image $I_t$, for each possible \newcomment{object pose and shape} $\S_j$ and vantage point $\g_t$, we can test the presence of at least one instance of each class $\M_k$ within. 
Note that the visibility function is implicit in the map $\pi$. If an object is not visible, its likelihood given the image $I_t$ is constant/uniform. Note that this depends on the global layout of the scene, since the map $\pi$ must take into account occlusions, so objects cannot be considered independently. 

\subsection{Dependencies and Co-visibility}
\label{sect-dependencies}

Computing the likelihood of an object being present in the scene requires ascertaining whether it is visible in the image, which in turn depends on all other objects, so the scene has to be modeled holistically rather than as an independent collection of objects. In addition, the presence of certain objects, and their configuration, affects the probability that other objects that are not visible be present.\footnote{For instance, seeing a keyboard and a monitor on a desk affects the probability that there is a mouse in the scene, even if we cannot see it at present. Their relative pose also informs the vantage point that would most reduce the uncertainty on the presence of the mouse.} 

To capture these dependencies, we note that the geometric representation $p(\g_t, \X | \y^t)$ can be used to provide a joint distribution on the position of all objects and cameras $p(\g^t, \X | \y^t)$, which yields {\em co-visibility} information, specifically the probability of each point in $\X$ being visible by any camera in $\g^t$. It is, however, of no use in determining visibility of objects, since it contains no topological information: We do not know if the space between two points is empty, or occupied by an object void of salient photometric features.
To enable visibility computation, we can use the point cloud together with the images to compute the {\em dense shape} of objects in a maximum-likelihood sense: $\hat \S_j = \arg\max p(\S_j | \g^t, \X, \y^t)$ using generic regularizers. This can be done but not at the level of accuracy and efficiency needed for live operation. An alternative is to approximate the shape of objects with a parametric family, for instance cuboids or ellipsoids, and compute visibility accordingly, also leveraging the co-visibility graph computed as a corollary from the SLAM system and priors on the size and aspect ratios of objects. 
To this end, we approximate
\be
\hspace{-.2cm}\hat p_{\g, \X}(\zi^j | \y^t) \doteq p(\zi^j | \y^t, \g_t, \X) \simeq \prod_j p(\zi_j | \y^t, \g_t, \X, \zi^{-j})
\ee
\comment{????? WHERE IS $o$???}
where $\zi^{-j}$ indicates all objects but $\z_j$. Each factor \newcomment{$p(\S_j, \M_j | \y^t, \g_t, \X, \zi^{-j})$} is then expanded as the product
\newcomment{
\be
 \underbrace{p(\S_j| \M_j, \y^t, \g_t, \X, \S^{-j})}_{\rm EKF} \underbrace{P(\M_j | \y^t, \g_t, \X, \M^{-j} )}_{\rm PMF} 
\label{eq-update}
\ee
}
where PMF indicates a probability mass filter; this effectively yields a bank of class-conditional EKFs. These provide samples from $\hat p(\zi | \y^t)$ in the right-hand side of \eqref{eq-approx}, that are scored with the CNN to update the posterior. \cut{Note that visibility is accounted for in the likelihood model of the EKF that, based on the sparse representation of the scene $\X$ and the shape $\S^{-j}$ and pose $\X^{-j}$ of other objects within. }

\cut{\subsection{Scene context} 
\label{sect-prior}
\begin{figure}[htb]
\begin{center}
\cut{\includegraphics[height=.8in]{context1.png}
\includegraphics[height=.8in]{context2.png}}
\end{center}
\caption{\sl  Effect of context: Imposing a scene prior that encodes co-occurrence helps reduce the likelihood of motorbike, aeroplane (left), and retrieve dining table and chair (right). The prior is learned from Pascal VOC2012.} \label{fig-context}
\end{figure}
\newcommand{\aeq}[1]{\begin{align} #1 \end{align}}
\newcommand{\aeqs}[1]{\begin{align*} #1 \end{align*}}
\newcommand{\beq}[1]{\begin{equation}#1\end{equation}}
\newcommand{\beqs}[1]{\begin{equation*}#1\end{equation*}}
\newcommand{\trm}[1]{\ensuremath \textrm{#1}}
\renewcommand{\d}{\delta}
\renewcommand{\a}{\alpha}
\newcommand{\reals}{\mathbb{R}}
\providecommand{\ind}{{\bf 1}}
\providecommand\f[2]{\ensuremath \frac{#1}{#2}}
\newcommand{\abs}[1]{\ensuremath \left| #1 \right|}
Geometric and topological context is already encoded in the representation, but we also wish to capture co-occurrence and scene-dependent category priors. To this end, we learn a hierarchical (Dirichlet) prior over co-occurrences of objects in the scene  using the Pascal VOC2012 dataset \cite{pascal-voc-2012} in a similar fashion as \cite{choi2010exploiting}. The posterior of objects' state is interpreted as the likelihood of the presence of a class; we use a message passing scheme computed using a tensor-train decomposition to compute the marginals of the model. Fig. \ref{fig-context} shows anecdotal examples of cases where the prior helps reject spurious classes (motorbike, plane) and help detect context-appropriate ones (table, bottle).
In practice, enforcing this prior is done by matrix multiplication of the class posterior estimates computed at each time, which rebalances the classes based on co-occurrence statistics in the training set.
}

\begin{table*}
\begin{center}
{\footnotesize
\begin{tabular}{|c|c|c|c|c|c|c|c|c|c|c|}
\hline
& Position error & \multicolumn{3}{c|}{$<0.5$~m} & \multicolumn{3}{c|}{$< 1$~m} & \multicolumn{3}{c|}{ $< 1.5$~m} \\
\hline
Orientation error & method & \#TP & Precision & Recall & \#TP & Precision & Recall & \#TP & Precision & Recall \\
\hline
\multirow{3}{*}{$ < 30^{\circ}$}
& {\tt Ours-FNL} &  \bf{150} & \bf{0.14} & \bf{0.10} & \bf{355} & \bf{0.34} & \bf{0.24} & \bf{513 }& \bf{0.49} & \bf{0.35} \\ \cline{2-11}
& {\tt Ours-INST} &  135 & 0.13 & 0.09 & 270 & 0.26 & 0.18 & 368 & 0.35 & 0.25 \\ \cline{2-11}
& {\tt SubCNN} &  99 & 0.10 & 0.07 & 254 & 0.26 & 0.17 & 376 & 0.38 & 0.26 \\
\hline\hline
\multirow{2}{*}{$ < 45^{\circ}$}
& {\tt Ours-FNL} &  \bf{157} & \bf{0.15} & \bf{0.11} & \bf{367} & \bf{0.35} & \bf{0.25} & \bf{533 }& \bf{0.50} & \bf{0.36} \\ \cline{2-11}
& {\tt Ours-INST} &  141 & 0.13 & 0.10 & 283 & 0.27 & 0.19 & 388 & 0.37 & 0.26 \\ \cline{2-11}
& {\tt SubCNN} &  99 & 0.10 & 0.07 & 257 & 0.26 & 0.17 & 383 & 0.38 & 0.26 \\
\hline\hline
\multirow{2}{*}{$ - $ }
& {\tt Ours-FNL} &  \bf{169} & \bf{0.16} & \bf{0.11} & \bf{425} & \bf{0.40} & \bf{0.29} & \bf{618 }& \bf{0.58} & \bf{0.42} \\ \cline{2-11}
& {\tt Ours-INST} &  149 & 0.14 & 0.10 & 320 & 0.30 & 0.22 & 450 & 0.43 & 0.31 \\ \cline{2-11}
& {\tt SubCNN} &  104 & 0.10 & 0.07 & 272 & 0.27 & 0.18 & 409 & 0.41 & 0.28 \\
\hline
\end{tabular}}
\end{center}
\caption{{\sl Quantitative evaluation on KITTI and comparison with SubCNN \cite{subcnn}}. \change{The number of true positives having positional error (row), and angular error (column) less than a threshold is shown, along with Precision and Recall. Scores are aggregated across all $3501$ ground-truth labeled frames in the dataset, with $498$ annotated objects. The last $3$ rows discard orientation error.}}
\label{tab-kitti-comp}
\end{table*}

\section{Implementation Details}
\label{sec:implementation}

\begin{figure}[t]
\begin{center}
\includegraphics[width=.84\columnwidth]{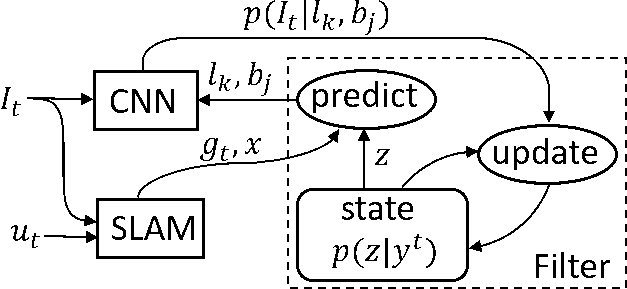}
\vspace{-.2cm}
\end{center}
 \caption{\change{{\sl System Flow Chart.}}}
\label{fig-flowchart}
\end{figure}

We have implemented two renditions of the above program: One operating in real-time and demonstrated live in June 2016 \cite{cvpr-demo-anonymous}. The other operating off-line and used for the experiments reported in Sect. \ref{sec:experiments}. \change{Fig.~\ref{fig-flowchart} sketches the system flow chart.}

In both cases, we have taken some shortcuts to improve the efficiency of the approximation of the likelihood function implemented by a CNN. Also, the semantic filter needs initialization and data association, which requires some heuristics to be computationally viable. We describe such heuristics in order.

\paragraph{Visual Odometry and Baseline 2D CNN}
We use robust SLAM implemented from \cite{tsotsosCS15} to acquire sparse point clouds and camera pose $x, g_t$ at each $t$. This occurs in $10-20$ms per VGA frame. \finalchange{For the quantitative evaluation on KITTI, we use \cite{mur2015orb} as the underlying localization pipeline.} For our real-time system, we use YOLO \cite{redmon2016you} as a baseline method to compute object likelihoods in $150-200$ms, whereas in the off-line system we use SubCNN \cite{subcnn}. In either case, the result is, for each given window, a positive score for each class $k$, read out from the penultimate layer. These are used both to compute the likelihood, and to generate proposals for initialization as discussed later. 
\vspace{-.4cm}
\paragraph{Filter Organization}
Each object is represented by a PMF filter over class labels and $K$ class-conditional EKFs, one for each class \eqref{eq-update}. Thus each object is represented by a mixture of $K$ EKFs, some of which pruned as we describe later. Each maintains a posterior estimate of position, scale and orientation relative to gravity. The state predicts the projection of (each of the $K$ instances of) each object onto the image plane, where the CNN evaluates the likelihood. For some object classes, we use a shape prior, enforced as a pseudo-measurement with uncertainty manually tuned to the expected class-variability. For instance, people are parallelepipeds of $1m^3$ expected volume with an anisotropic covariance along coordinate axes in the range of few decimeters, whereas couches have significantly more uncertainty. 
\vspace{-.4cm}
\paragraph{Data Association}
To avoid running the baseline CNN multiple times on overlapping regions (each object is represented by multiple, often very similar, regions, one per each current class hypothesis), we do not query the CNN sequentially for each prediction. Instead, we run the CNN once, with lax threshold so as to obtain a large number of (low-confidence) regions. While this is efficient, it does create a data association problem, as we must attribute (possibly multiple) image regions to each (of multiple) object hypotheses, each of which has multiple possible class labels \cite{atanasov2014semantic}. \change{We avoid explicit data association by opting simple heuristics instead: first we generate predictions from the filter; then occluded objects are excluded from likelihood evaluation. For all others, we generate four-tuple coordinates of the bounding box, as a $4$-dimensional Gaussian given the projection of the current state. This is a sloppy prediction, for the image of a parallelepiped is in general not an axis-aligned rectangle on the image. Nevertheless, we use this for scoring the use of the likelihood produced by the CNN for each predicted class. A (class-dependent) threshold is used to decide if the bounding box should be used to update the object. Bounding boxes with lower likelihood are given small weights in the filter update.} This requires accurate initialization, which we will describe below. The silver lining is that inter-frame motion is usually small, so data association proceeds smoothly, unless multiple instances of the same object class are present nearby and partially occlude each other.
\vspace{-.4cm} 
\paragraph{Initialization} Putative 2D CNN detections not associated to any object are used as (bottom-up) proposals for initialization. The new object is positioned at the weighted centroid of the sparse points whose projections lie within the detection region. The weight at center is the largest and decreases exponentially outwards. Orientation is initialized as the ``azimuth'' from SubCNN, rotated according to camera pose and gravity. Given the position and orientation, scale is optimized by minimizing the reprojection error. \newcut{A class-conditional aspect ratio prior can be used again as a pseudo-measurement.}
\vspace{-.4cm}
\paragraph{Merge} Objects are assumed to be simply-connected and compact, so two objects cannot occupy the same space. Yet, their projected bounding boxes can overlap. If multiple instances from the same object are detected, initialized and propagated, they will eventually merge when their overlap in space is sufficiently large. Only objects from the same class are allowed to merge as different classes may appear co-located and intersecting in their sloppy parallelepipedal shape model, \eg, a chair under a table.
\vspace{-.4cm}
\paragraph{Termination}  Each object maintains a probability over $K$ classes, each associated with a class-conditional filter. If one of the classes becomes dominant (maximum probability above a threshold), all other filters will be eliminated to save computational cost. Most objects converge to one or two classes (\eg, chair, couch) within few iterations. Objects that disappear from view are retained in the state (short-term memory), and if not seen for a sufficiently long time, they are stored in long-term memory (``semantic map'') for when they will be seen again.

\change{There are more implementation details that can be described in the space available. For this reason, we make our implementation publicly available at \cite{Appendix}.}

\begin{figure}[t]
\begin{center}
\subfigure{\includegraphics[width=.24\columnwidth]{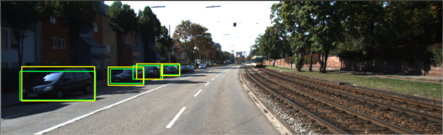}}
\subfigure{\includegraphics[width=.24\columnwidth]{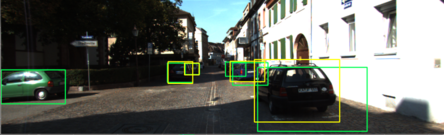}}
\subfigure{\includegraphics[width=.24\columnwidth]{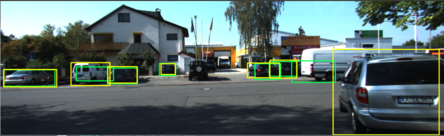}}
\subfigure{\includegraphics[width=.24\columnwidth]{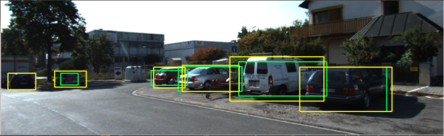}}\vspace{-.3cm}\\
\subfigure{\includegraphics[width=.24\columnwidth]{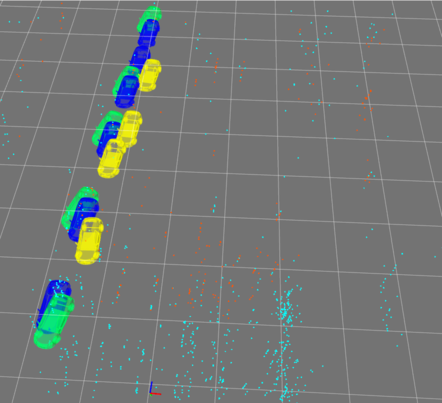}}
\subfigure{\includegraphics[width=.24\columnwidth]{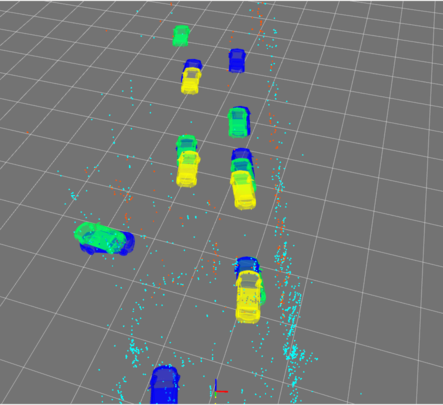}}
\subfigure{\includegraphics[width=.24\columnwidth]{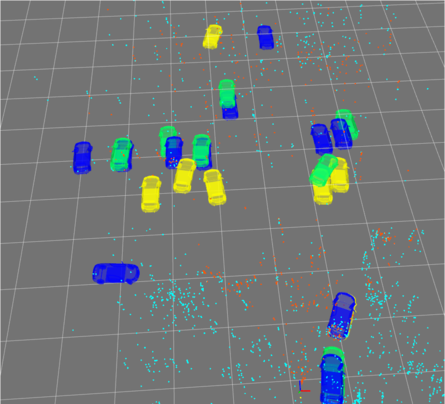}}
\subfigure{\includegraphics[width=.24\columnwidth]{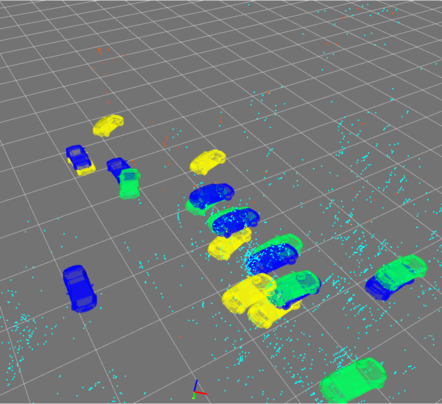}}\vspace{-.3cm}
\end{center}
   \caption{{\sl Qualitative comparison with SubCNN.} \change{Top: Images with back-projected objects from our method (Green), the same with SubCNN (Yellow). Bottom: top-view of the corresponding portion of the scene. Ground truth is shown in Blue.}}
\label{fig-kitti-exmp}
\vspace{-.3cm}
\end{figure}

\begin{figure*}[t]
\begin{center}
\subfigure{\label{fig-update-loc-a}\includegraphics[width=.16\textwidth]{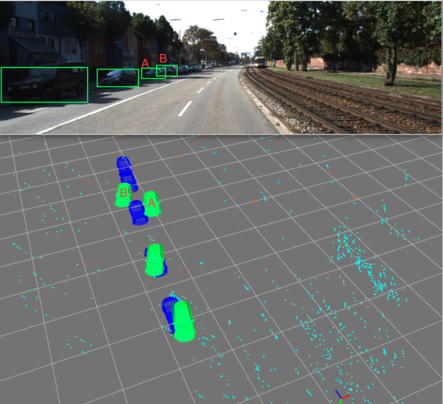}}
\subfigure{\label{fig-update-loc-d}\includegraphics[width=.16\textwidth]{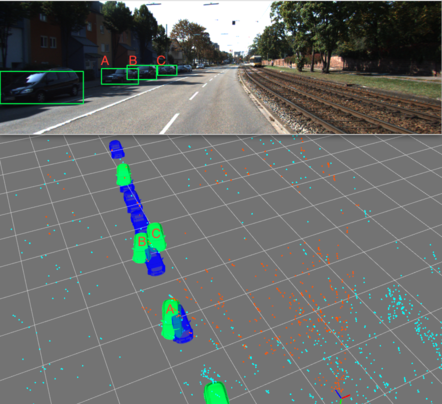}}
\subfigure{\label{fig-update-loc-e}\includegraphics[width=.16\textwidth]{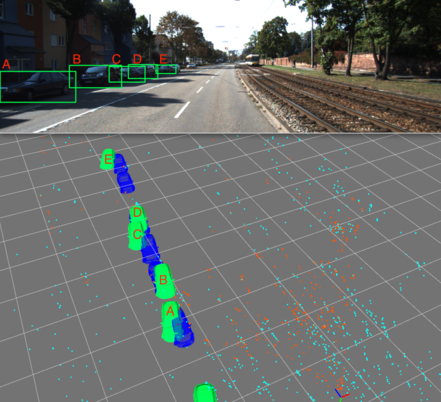}}
\subfigure{\label{fig-update-loc-f}\includegraphics[width=.16\textwidth]{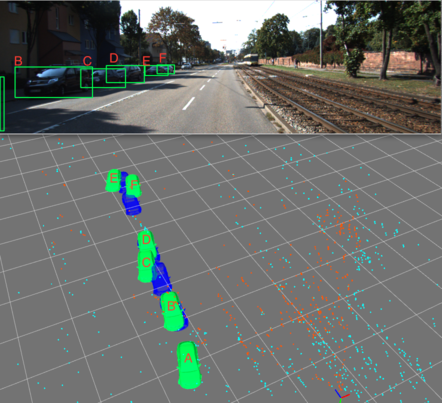}}
\subfigure{\label{fig-update-loc-h}\includegraphics[width=.16\textwidth]{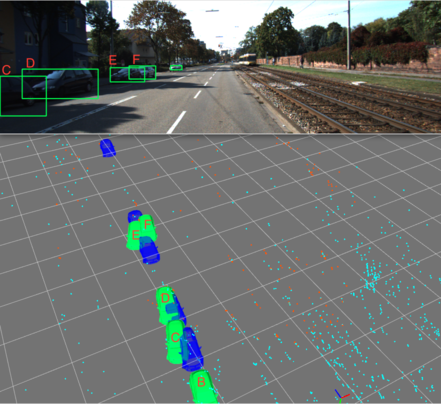}}
\subfigure{\label{fig-update-loc-j}\includegraphics[width=.16\textwidth]{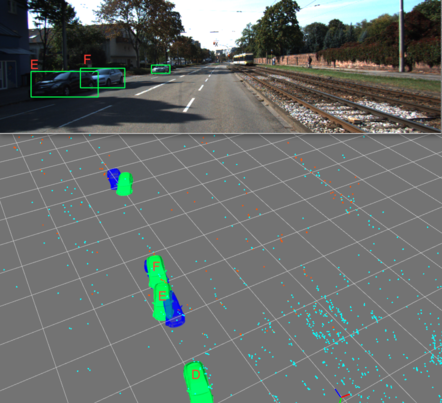}}\vspace{-.3cm}
\end{center}
   \caption{{\sl Evolution of the state (Green) against ground-truth annotation (Blue) (best viewed at $5\times$, images shown at the top for ease of reference).} When first seen (Leftmost) cars `A' and `B' are estimated to be side-by-side; after a few frames, however, `A' and `B' fall into place, but a new car `C' appears to flank `B'. As time goes by, `C' too falls into place, as new cars appear, `D', `E', `F.' The error in pose (position and orientation) relative to ground truth can be appreciated qualitatively. Quantitative results are shown in Table \ref{tab-kitti-comp}.}
\label{fig-update-loc}
\vspace{-.3cm}
\end{figure*}


\section{Experiments}
\label{sec:experiments}

\subsection{Quantitative Results}
\newcut{We compare our results with SubCNN \cite{subcnn} which produces, for each image, multiple image-bounding boxes, each associated with a 3DVP \cite{3dvp} learned from a separate training set. Meta-data, \eg, shape (represented as voxelized CAD models) and pose, can be retrieved directly from the associated 3DVPs. Given camera intrinsics, a bounding box in 2D and meta-data can be used to back-project objects into the scene, by minimizing the reprojection error.} 

As explained in Sec.~\ref{sect-relatedwork}, we choose SubCNN \cite{subcnn} as the paragon, even though it is based on a single image, because it is the top performer for 3D recognition in KITTI among non-anonymous and reproducible ones, in particular it dominates \cite{chen2016monocular}. Being single-image based, SubCNN returns different results in each frame, therefore naturally at a disadvantage. To make the comparison fair, one would have to average or integrate detections for each object across all frames when it is visible. However, SubCNN does not provide data association, making direct comparison challenging. To make comparison as fair as possible, without developing an alternate aggregation method for SubCNN, we compare it to our algorithm on a frame-by-frame basis. Specifically, for each frame, we transfer the ground truth to the camera frame, and remove occluded objects. Then we can compare detections from SubCNN to our point estimate (conditional mean) computed causally by the filter at the current time. We call this method {\tt Ours-INST}. On the other hand, we can benefit from aggregating temporal information for as long as possible, so we also report results based on the point-estimate of the filter state at the last time instant when each object is seen. The estimate is then mapped back to the current frame, which we call {\tt Ours-FNL}. To the best of our knowledge, there are no known methods for 3D recognition that causally update posterior estimates of object identity/presence and geometric attributes, and even naive temporal averaging of a method like \cite{subcnn} is not straightforward because of the absence of data association across different frames. This is precisely what motivates us.

\vspace{-.2cm}
\subsubsection{Dataset}
There are many datasets for image-based object detection \cite{pascal, imagenet} which provide 2D ground truth. There are also 3D object detection datasets \cite{xiao2013sun3d}, most using extra sensor data, \eg, depth from a structured-light sensor. None provide inertial measurements, except KITTI \cite{kitti}, {whose object detection benchmark contains $7181$ images, from which we exclude $3682$ frames used for SubCNN training \cite{subcnn}, leaving us a validation set of $3799$ frames. \change{We then find $10$ videos which cover most of the validation set. After removing moving objects, $498$ objects are observed $18468$ times at $3501$ instants, which is the same order of magnitude of the 2D validation set.}} 

\vspace{-.2cm}
\subsubsection{Evaluation Metrics}

KITTI provides ground-truth object {\em tracklets} we use to define true positives, miss detections and false alarms. A {\em true positive} is the nearest detection of a ground truth object within a specified error threshold in both position and orientation (Table~\ref{tab-kitti-comp}). A {\em miss} occurs if there is no detection within the threshold. A {\em false alarm} occurs when an object is detected despite no true object being within the threshold in distance and orientation. {\em Precision} is the fraction of true positives over all detections, and {\em Recall} is the percentage of detected instances among all true objects.

\vspace{-.2cm}
\subsubsection{Benchmark Comparison} 
Table~\ref{tab-kitti-comp} shows result on the KITTI dataset, averaged over all sequences. On average, {\tt Ours-INST} already outperforms {\tt SubCNN} even if our initialization can be rather inaccurate. \change{Note that our method requires evidence to be accumulated over time before claiming the existence of an object in the scene, so {\tt Ours-INST} is penalized heavily in the first few frames when a new object is spotted.}  {\tt Ours-FNL} further improves the results by a large margin. Fig.~\ref{fig-update-loc} shows how our method refines the state over time. Visual comparison is shown in Fig.~\ref{fig-kitti-exmp} for ground truth (Blue), {\tt Ours-FNL} (Green) and {\tt SubCNN} (Yellow). \newcut{More illustrative examples are in the Sup. Mat.}


\subsection{Class-specific Priors}
\label{sec:scale}
\begin{figure}[t]
\begin{center}
\subfigure{\label{fig-toy-car-b}\includegraphics[width=.48\columnwidth]{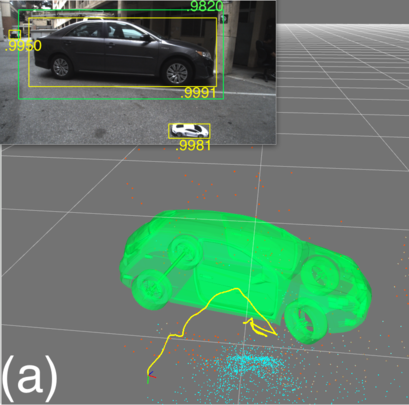}}
\subfigure{\label{fig-toy-car-c}\includegraphics[width=.48\columnwidth]{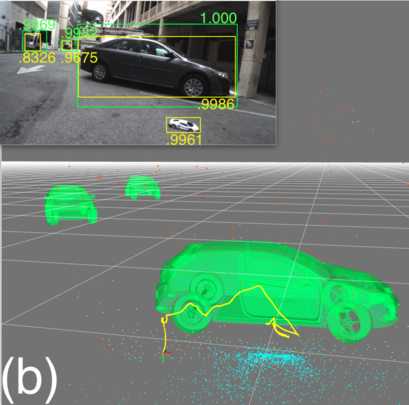}}\vspace{-.3cm}
\end{center}
   \caption{{\sl Class-specific scale prior.} \subref{fig-toy-car-b}: A real car is detected by our system, unlike the toy car, despite both scoring high likelihood and therefore being detected by an image-based system (Yellow). As time goes by, the confidence on the real car increases (best viewed at $5\times$) \subref{fig-toy-car-c}. \change{See online video at \cite{Appendix}.}}
\label{fig-toy-car}
\end{figure}

Objects have characteristic scales, which are lost in perspective projection but inferable with an inertial sensor. We impose a class-dependent prior on size and shape (\eg, volume, aspect ratios). In Fig.~\ref{fig-toy-car}, a toy car is detected as a car by an image-based detector (Yellow), but rejected by our system as inconsistent with the scale prior (Green). Fig.~\ref{fig-toy-car-c} shows two background cars in the far field, whose images are smaller than the toy car, yet they are detected correctly, whereas the toy car is rejected.

\subsection{Occlusion and Memory}
\label{sec:occlusion}
\begin{figure}[t]
\begin{center}
\subfigure{\label{fig-occ-a}\includegraphics[width=.32\columnwidth]{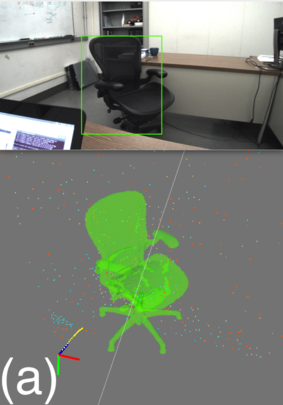}}
\subfigure{\label{fig-occ-b}\includegraphics[width=.32\columnwidth]{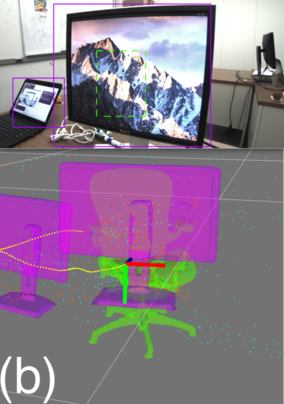}}
\subfigure{\label{fig-occ-c}\includegraphics[width=.32\columnwidth]{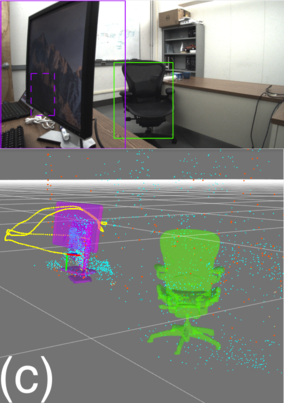}}\vspace{-.3cm}
\end{center}
   \caption{{\sl Occlusion management and short-term memory.} \subref{fig-occ-a}: A chair is detected and later becomes occluded by the monitor \subref{fig-occ-b}. Its projection onto the image is shown in dashed lines, indicating occlusion. The model allows prediction of dis-occlusion \subref{fig-occ-c} which allows resuming update when the chair comes back into view. \change{See online video at \cite{Appendix}.}}
\label{fig-occ}
\vspace{-.2cm}
\end{figure}

Our system represents objects in the state even while they are not visible, or detected by an image-based detector. This allows predicting the re-appearance of objects in future frames, and to resume update if new evidences appear. Fig.~\ref{fig-occ} shows a chair first detected and then occluded by a monitor, later reappearing. The system predicts the chair to be completely occluded, and therefore does not use the image to update the chair, but resumes doing so when it reappears, by which time it is known to be the {\em same} chair that was previously seen (re-detection). 
In Sect.~\ref{sect-kitti00}, we show the same phenomenon in a large-scale driving sequence. 

\subsection{Large-scale Driving Sequences}
\label{sect-kitti00}
Fig.~\ref{fig-teaser} and \change{online video at \cite{Appendix}} show our results on a $3.7$km-long sequence from KITTI. It contains hundreds of cars along the route. Once recognized as a car, we replace the bounding box with a CAD model of similar car, aligned with the pose estimate from the filter, in a manner similar to \cite{salas2013slam++}, that however uses RGB-D data. In this sequence, we can also see cars on different streets ``through walls'' if they have been previously detected, which can help navigation.

\subsection{Indoor Sequences}

\begin{figure}[t]
\begin{center}
\subfigure{\includegraphics[width=.30\columnwidth]{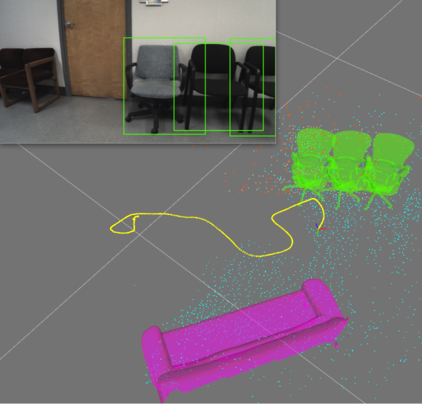}}
\subfigure{\includegraphics[width=.30\columnwidth]{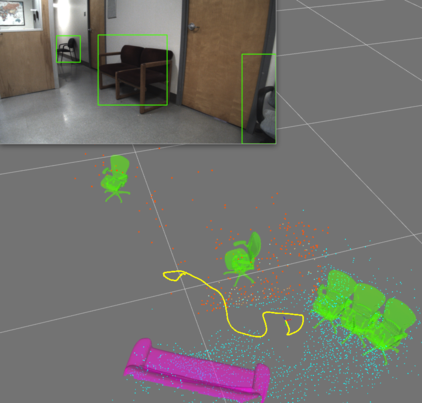}}
\subfigure{\includegraphics[width=.30\columnwidth]{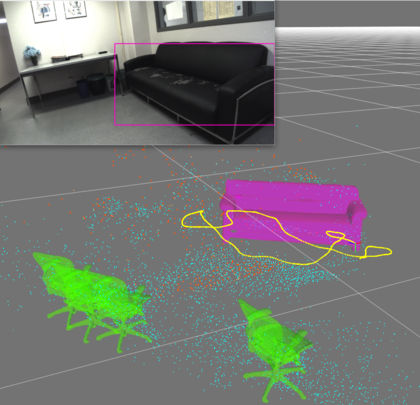}}\vspace{-.3cm}\\
\subfigure{\includegraphics[height=.25\columnwidth]{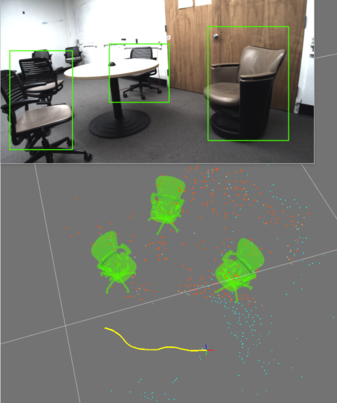}}
\subfigure{\includegraphics[height=.25\columnwidth]{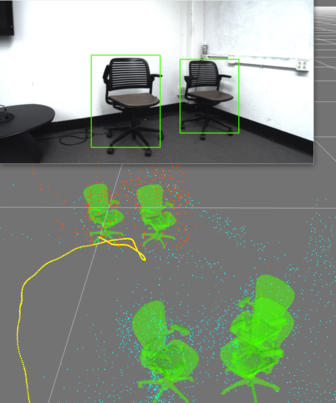}}
\subfigure{\includegraphics[height=.25\columnwidth]{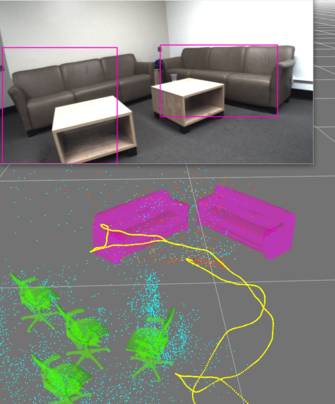}}
\subfigure{\includegraphics[height=.25\columnwidth]{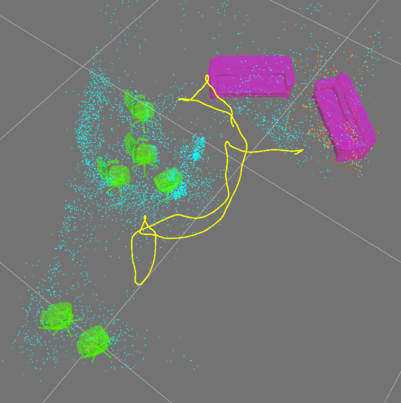}}\vspace{-.3cm}
\end{center}
   \caption{{\sl Indoor sequences.} Top: An office area. Bottom: A Lounge area. \change{Both videos are available at \cite{Appendix}.} }
\label{fig-indoor}
\vspace{-.2cm}
\end{figure}

We have tested our system live in a public demo \cite{cvpr-demo-anonymous}, operating in real time in cluttered environments with people, chairs, tables, monitors and the like. Representative examples are shown for simpler scenes, for illustrative purposes, in Fig.~\ref{fig-indoor}, where again CAD models of objects are rendered once detected, a' la \cite{salas2013slam++}. Our system does not produce exact orientation estimates, as seen in Fig.~\ref{fig-indoor}, so there is plenty of room for improvement.

\section{Discussion}
\label{sec:discussion}

Inertial sensors are in every modern phone, tablet, car, even many toys, all devices embedded in physical space and occasionally in need to interact with it. It makes sense to exploit inertials, along with visual sensors, to help detecting objects that exist in 3D physical space, and have characteristic shape and size, in addition to appearance. We have recorded tremendous progress in object detection in recent years, if by object one means a group of pixels in an image. Here we leverage such progress to design a detector that follows the prescriptions (a)-(e) indicated in the introduction. 

We start by defining a representation as a minimal sufficient invariant statistic of object attributes, in line with \cite{soattoC16ICLR}. We then marginalize on camera Euclidean pose -- which allows us to enforce priors on the class-specific scale of objects -- and update the measure by a Bayesian filter, where a CNN is in charge of computing the likelihood function. 

We note that a minimal sufficient invariant for localization is an attributed point cloud, and therefore there is no need to deploy the machineries of Deep Learning to determine camera pose (Deep Learning could still be used to infer the attributes at points, which are used for correspondence). Instead, we use an Extended Kalman Filter, conditioned on which the update for object attributes can be performed by a Mixture-of-Kalman filter.

The result is a system whereby objects do not flicker in-and-out of existence, our confidence in their presence grows with accrued evidence, we know of their presence even if temporarily occluded, we can predict when they will be seen, and we can enforce known scale priors to reject spurious hypotheses from the bottom-up proposal mechanism. 

We have made stringent and admittedly restrictive assumptions in order to keep our model viable for real-time inference. One could certainly relax some of these assumptions and obtain more general models, but forgo the ability to operate in real time. 

The main limitation of our system is its restriction to static objects. While in theory the framework is general, the geometry of moving and deforming objects is not represented, and therefore their attributes remain limited to what can be inferred in the image. Also, our representation of objects' shape is rather rudimentary, and as a result visibility computation rather fragile. These are all areas prime for further future development.

\change{Our datasets, consisting of monocular imaging sequences with time-stamped inertial measurements, are available at \cite{Appendix}, along with our system implementation.}

\section*{Acknowledgments}
\change{Research sponsored by ARO W911NF-15-1-0564/66731-CS, ONR N00014-17-1-2072, AFOSR FA9550-15-1-0229.}


\end{document}